\documentclass[runningheads]{llncs}
\usepackage{graphicx,wrapfig}
\usepackage{caption}
\usepackage{subcaption}
\usepackage{wrapfig}
\usepackage{geometry}
\usepackage{amsmath}
\usepackage{hyperref}
\usepackage{adjustbox}
\begin{document}
\title{Preventing Deterioration of Classification Accuracy in Predictive Coding Networks}
%
%

\author{Paul F Kinghorn\inst{1}, Beren Millidge\inst{2}\and
Christopher L Buckley\inst{3}}

\authorrunning{}
\institute{School of Engineering and Informatics,
University of Sussex, \email{p.kinghorn@sussex.ac.uk} 
\and MRC Brain Networks Dynamics Unit,
University of Oxford, \email{beren@millidge.name}
\and School of Engineering and Informatics,
University of Sussex, \email{c.l.buckley@sussex.ac.uk}
}

\maketitle              
\begin{abstract}
Predictive Coding Networks (PCNs) aim to learn a generative model of the world. Given observations, this generative model can then be inverted to infer the causes of those observations. However, when training PCNs, a noticeable pathology is often observed where inference accuracy peaks and then declines with further training. This cannot be explained by overfitting since both training and test accuracy decrease simultaneously. Here we provide a thorough investigation of this phenomenon and show that it is caused by an imbalance between the speeds at which the various layers of the PCN converge. We demonstrate that this can be prevented by regularising the weight matrices at each layer: by restricting the relative size of matrix singular values, we allow the weight matrix to change but restrict the overall impact which a layer can have on its neighbours. We also demonstrate that a similar effect can be achieved through a more biologically plausible and simple scheme of just capping the weights.

\end{abstract}

\section{Introduction}
Predictive Coding (PC) is an increasingly influential theory in computational neuroscience, based on the hypothesis that the primary objective of the cortex is to minimize prediction error ~\cite{ClarkAndy2013WnPb,millidge2021predictive,seth2014cybernetic}. Prediction error represents the mismatch between predicted and actual observations. The concepts behind PC go back to Helmholtz's unconscious inference and the ideas of Kant~\cite{millidge2021predictive}. There are also more recent roots in both machine learning~\cite{beal2003variational,dayan1995helmholtz} and neuroscience~\cite{mumford1992computational,rao1999predictive} which were then unified by Friston in a series of papers around 15 years ago~\cite{friston2003learning,friston2005theory,friston2008hierarchical}. In order to generate predictions, the brain instantiates a generative model of the world, producing  sensory observations from latent variables. Typically, PC is assumed to be implemented in hierarchies, with each layer sending predictions down to the layer below it, although recent work has demonstrated its applicability in arbitrary graphs~\cite{salvatori2022learning}.  
A key advantage of PC over backpropagation is that it requires only local updates. Despite this, recent work has shown that it can approximate backpropagation~\cite{song2020can,millidge2022predictive}, and it is an active area of research in both machine learning and neuroscience. 

A Predictive Coding Network (PCN) can be viewed as an example of an Energy Based Model (EBM), with the minimization of errors equating to  minimization of the network's energy \cite{friston2005theory}. Minimizing energy by updating network node values corresponds to inference and perception, whereas reducing energy by updating the weights of the network corresponds to learning and improving the model of the world. The theory has close links with the concept of the Bayesian Brain~\cite{seth2014cybernetic,knill2004bayesian,doya2007bayesian} - the process of perception is implemented by setting a prior at the top layer being sent down the layers and then errors being sent back up the layers to create a new posterior given observations at the bottom of the network. This is done iteratively until the posterior ``percept'' at the top of the network and the incoming data at the bottom are in equilibrium which is when the energy of the network is minimized. 

As the generative model is learned and the weights of the network are updated, the ability of the network to infer the correct latent variable (or label) should, in theory, improve.  However, we have observed that, once an optimal amount of training has occurred, PCNs appear to then deteriorate in classification performance, with inference having to be run for increasingly many iterations to maintain a given level of performance. We are aware of only passing mentions of this issue in the literature~\cite{kinghorn2021habitual,tschantz2022hybrid}, although it is often discussed informally. This paper provides an in-depth investigation and diagnosis of the problem, determines the reasons for it and then implements some techniques which can be used to avoid it.  This allows us to stably train predictive coding networks for much longer numbers of epochs than previously possible without performance deterioration.

The remainder of this paper is set out as follows. Section 2 describes a typical PCN and gives a high level overview of the maths behind predictive coding. Section 3 analyses the reason for the degradation in performance, starting with a demonstration of the problem and then explains its causes at increasingly detailed levels of explanation, ultimately demonstrating that it is caused by a mismatch between the size of weights in different layers. Once we have identified this fundamental cause, Section 4 then demonstrates techniques for avoiding it such as weight regularisation or capping. Weight regularisation is a technique which is common in the world of machine learning to prevent overfitting~\cite{bishop2006pattern}. However, we will demonstrate that this is not the problem faced in PCNs - our solutions are not designed to necessarily keep weights small, but rather to ensure that the relative impact of different weight layers stays optimal.   

\section{Predictive Coding Networks}

\begin{wrapfigure}{r}{0.5\linewidth}
\vspace{-0.4cm}
 \captionsetup{justification=raggedright}
  \centering
  \includegraphics[width=.99\linewidth]{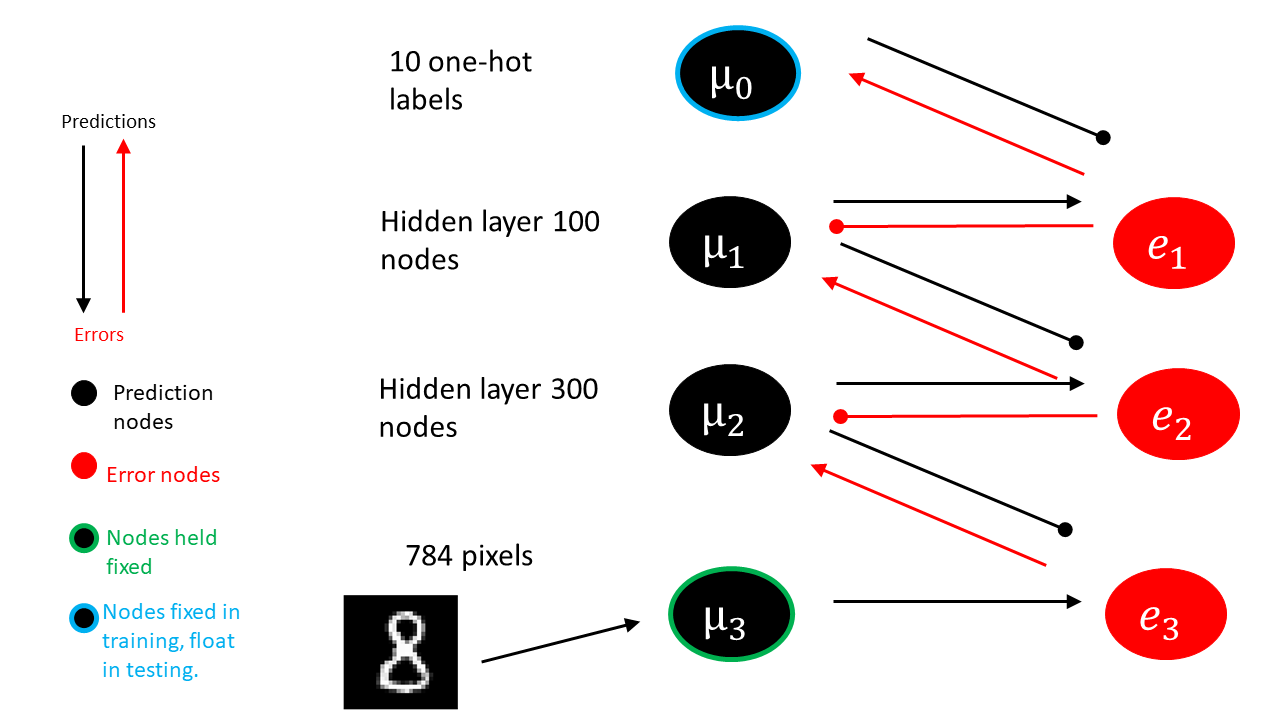}
\caption{Predictive Coding Network in test mode. The PCN learns to generate images from labels. 
At test time, the task of the network is to infer the correct label from a presented image. 
This is done by iteratively sending predictions down and errors back up the network, until the network reaches equilibrium. 
}
\label{network}
\vspace{-0.4cm} \end{wrapfigure}

PCNs can be trained to operate in two different manners. Using MNIST~\cite{lecun-mnisthandwrittendigit-2010} as an example, a ``generative'' PCN would generate images from labels in a single forward sweep, but then be faced with a difficult inversion task in order to infer a label from an image. Conversely, a ``discriminative'' PCN would be able to generate labels from images in a  single forward sweep but would have difficulty producing an image for a given label~\cite{millidge2021predictive}. Throughout this paper we use generative PCNs. Fig.~\ref{network} shows a typical PCN. 
In training, it learns a generative model which takes MNIST labels at the top of the network and generates MNIST images at the bottom. 
When testing the network's ability to classify MNIST images, an image is presented at the bottom of the network and the generative model is inverted using variational inference to infer a label.
Derivations of the maths involved can be found in~\cite{BuckleyChristopherL2017Tfep,bogacz2017tutorial,millidge2019combining,kinghorn2021habitual} and we do not derive the update equations here, but simply give a brief overview and present the update equations which will be relevant later.

The network has N layers, where the top layer is labelled as layer 0 and the bottom layer as layer N-1. \(\mu_{n}\) is a vector representing the node values of layer \(n\), \(\theta_{n}\) is a matrix giving the connection weights between layer \(n\) and layer \(n+1\), \(f\) is an elementwise non-linear function and \(\epsilon_{n+1} := \mu_{n+1}-f(\mu_{n}   \theta_n)\) is the difference between the value of layer \(n+1\) and the value predicted by layer \(n\). Note that, in this implementation, \((\mu_{n}   \theta_n)\) represents matrix multiplication of the node values and the weights, and therefore the prediction sent from layer \(n\) to layer \(n+1\) is a non-linear function applied elementwise to a linear combination of layer \(n\)'s node values. This is simply a specific instance of the more general case, where the prediction is produced by an arbitrary function \(f\) parameterised by \(\theta_n\), and is therefore given by \(f(\mu_{n} ,  \theta_n)\).     

Training the generative model of the network involves developing an auxiliary model (called the variational distribution) and minimizing the KL divergence between that auxiliary model and the true posterior. Variational free energy \(\mathcal{F}\) is a measure which is closely related to this divergence and under certain assumptions it can be shown that: 
\begin{equation}
\label{F}
\begin{split}
\mathcal{F}  
\approx \  \Big[ \ \sum_{n}^N -\frac{1}{2}  \epsilon_{n+1}^T \Sigma_{n+1} ^{-1} \epsilon_{n+1} - \ \frac{1}{2} log (2 \pi \ | \Sigma_{n+1} |) \ \Big]
\end{split}
\end{equation}
where \(\Sigma_n^{-1}\) is a term known as the precision, equal to the inverse of the variance of the nodes. 
It can also be shown that, in order to make the model a good fit for the data, it suffices to minimize \(\mathcal{F}\). 
Therefore, to train the model, batches of label/image combinations are presented to the network and \(\mathcal{F}\) is steadily reduced using the Expectation-Maximization approach~\cite{dempster1977maximum,mackay2003information}, which alternately applies gradient descent on \(\mathcal{F}\) with respect to node values (\(\mu\)) on a fast timescale and weight values (\(\theta\)) on a slower timescale. The process is repeated over multiple batches, with \(\mathcal{F}\) steadily decreasing and the network's generative model improving. The gradients can be easily derived from equation~\eqref{F}.  To update the nodes of the hidden layers, the gradient is given by:
\begin{equation}
\label{updates2}
\frac{d\mathcal{F}}{d\mu_n}=
\epsilon_{n+1} \  
\Sigma_{n+1}^{-1} \  \theta_n^T \ 
f'(\mu_n \theta_n)  - 
\epsilon_{n} \ 
\Sigma_{n}^{-1} 
\end{equation}

After the node values have been updated, \(\mathcal{F}\) is then further minimized by updating the weights using:

\begin{equation}
\label{updates1}
\frac{d\mathcal{F}}{d\theta_n} =
{\epsilon_{n+1} \ \Sigma_{n+1}^{-1}} \
\mu_n^T \ 
f'(\mu_n \theta_n) %
\end{equation}

Training is carried out in a supervised manner and involves presentation of images and their corresponding labels. Therefore the bottom and top layers of the network are not updated during training and are held fixed with an MNIST image and a one-hot representation of the image's MNIST label respectively. Only the hidden layer nodes are updated. During testing, the task is to infer a label for a presented image and therefore the nodes of the top layer are allowed to update. Using the gradient for node updates, gradient descent is run through a number of iterations, ideally until the nodes of the network are at equilibrium. The inferred label can then be read out from the top layer. Since this layer receives no predictions from above, its gradient is slightly different from the other layers and is truncated to: 
\begin{equation}
\label{updates3}
\frac{d\mathcal{F}}{d\mu_0}=
\epsilon_{1} \
\Sigma_{1}^{-1} \  \theta_0^T \
f'(\mu_0 \theta_0) 
\end{equation}

So far, we have described a supervised training regime. It is also possible to train in an unsupervised manner, in which case the top layer nodes are not held fixed during training and are updated in the same way as they are in testing. In this scenario, the top layer nodes will not be trained to converge to one-hot labels when inferring an image, but will still try to extract latent variables (see Appendix~\ref{appendix:tsne} for more discussion). 

It is also possible to derive update equations for the precisions \cite{friston2005theory,bogacz2017tutorial,millidge2021predictive}, but in our experiments we hold these fixed as the identity matrix and do not update them.
Details of the network and the gradient descent techniques used are given in Appendix~\ref{appendix:network_details}.

\section{Analysis}

\begin{figure}
\vspace{-0.4cm}
\centering

\begin{minipage}{.49\textwidth}
\centering
\includegraphics[width=.98\linewidth]{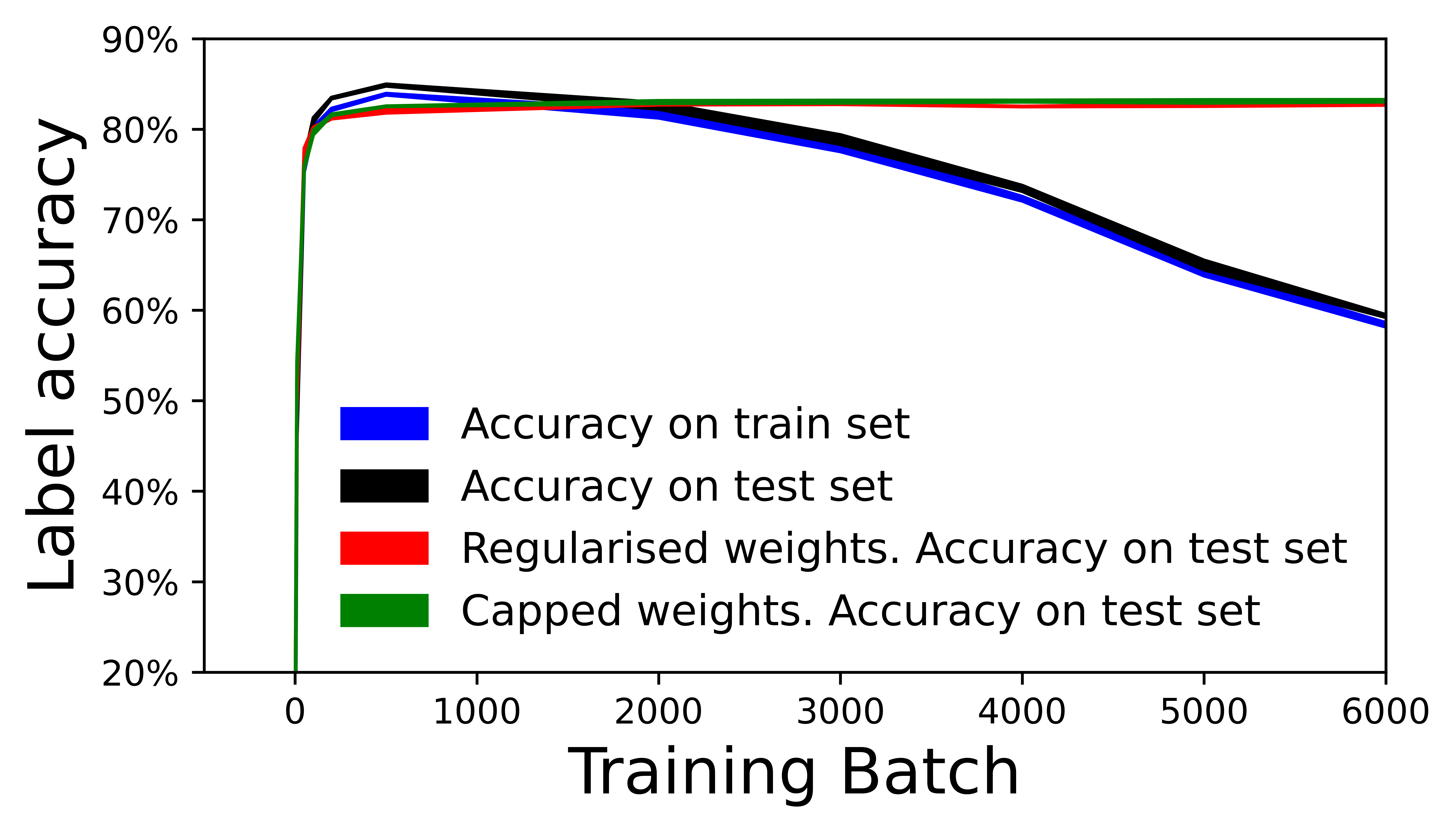}
\captionsetup{justification=raggedright}
\caption{\textbf{Deteriorating Inference Accuracy.} After training progresses beyond a certain stage, the ability of a PCN to infer the correct latent variable (``label'') decreases.  This is observed for both test and training set (black and blue lines) and therefore cannot be attributed to over-fitting. This paper demonstrates that we can prevent this issue by simply capping size of weights (green line) or by regularising the weights so that the mean weight size on each layer remains constant (red line). Each line shows average of 3 networks, with standard error shown.}
\label{diminishing_accy}
\end{minipage}
\hfill
\begin{minipage}{.49\textwidth}
\centering
\includegraphics[width=.98\linewidth]{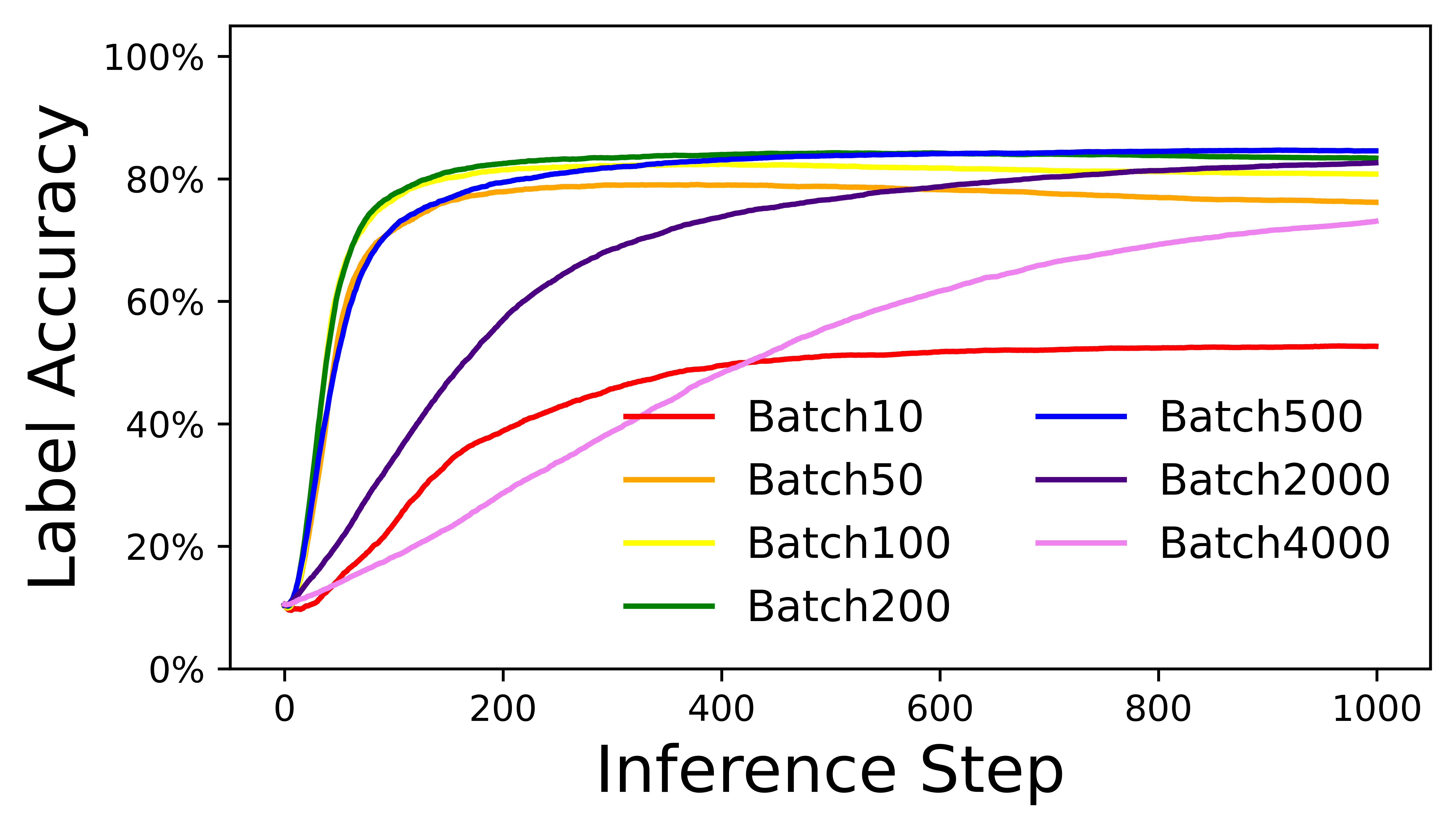}
\captionsetup{justification=raggedright}
\caption{\textbf{For different levels of training, development of label accuracy during iterative inference.} During the early stages of training (roughly batches 10 to 200), the  number of inference iterations required in testing reduces. However, after a certain amount of training, the  asymptotic accuracy no longer improves, and the time taken to reach that asymptote worsens. \newline This plot shows results from a single run - different runs vary slightly, but follow the same pattern.\newline
}
\label{inference_speed}
\end{minipage}

\vspace{-0.4cm}
\end{figure}

Fig.~\ref{diminishing_accy} demonstrates a problem which is encountered when training generative PCNs for inference/classification tasks. In order to assess how well the network infers labels of MNIST images, we train the network over a large number of mini-batches and regularly test the accuracy of label inference against a test dataset (black line), using 1000 iterations of inference. The network quickly improves with training, but accuracy then appears to deteriorate once training progresses beyond a certain point. 
The remainder of the paper demonstrates that this is caused by a mismatch between the way weights in different layers develop during training. We then show that simply capping the size of weights or regularising the weights so that the mean weight
size on each layer remains constant prevents the problem and stabilizes training.  

At any step in the  iterative inference, it is possible to take a one-hot read-out from the inferred labels at the top layer. We can therefore construct a trajectory showing how accuracy develops with number of inference iterations. Fig.~\ref{inference_speed} shows how this label accuracy develops for a network given different amounts of training. For example, the red line in the figure shows the trajectory of label accuracy for a network which has received 10 mini-batches of training.   The plot demonstrates that, as training progresses through the early training batches, test accuracy improves both in terms of the asymptotic value achieved and in terms of how quickly the inference process reaches that asymptote. But, after a certain amount of training the inference process slows down. At batch 200, asymptotic performance is achieved after approximately 100 iterations. But if the network is trained for 4000 batches, the accuracy is still improving after 1000 test iterations. It is important to note that, if the inference process were allowed to run longer, the same asymptotic value would be achieved. Thus, if we infer the label after a set number of test iterations, performance will seem to deteriorate as the network gets better trained. This explains why figure~\ref{diminishing_accy} appears to show a deterioration in network performance since, following common practice, we stopped inference after a fixed number of steps (1000).

\begin{figure}
 \captionsetup{justification=raggedright}
\centering

\begin{subfigure}{.325\textwidth}
  \centering
  \includegraphics[width=.995\linewidth]{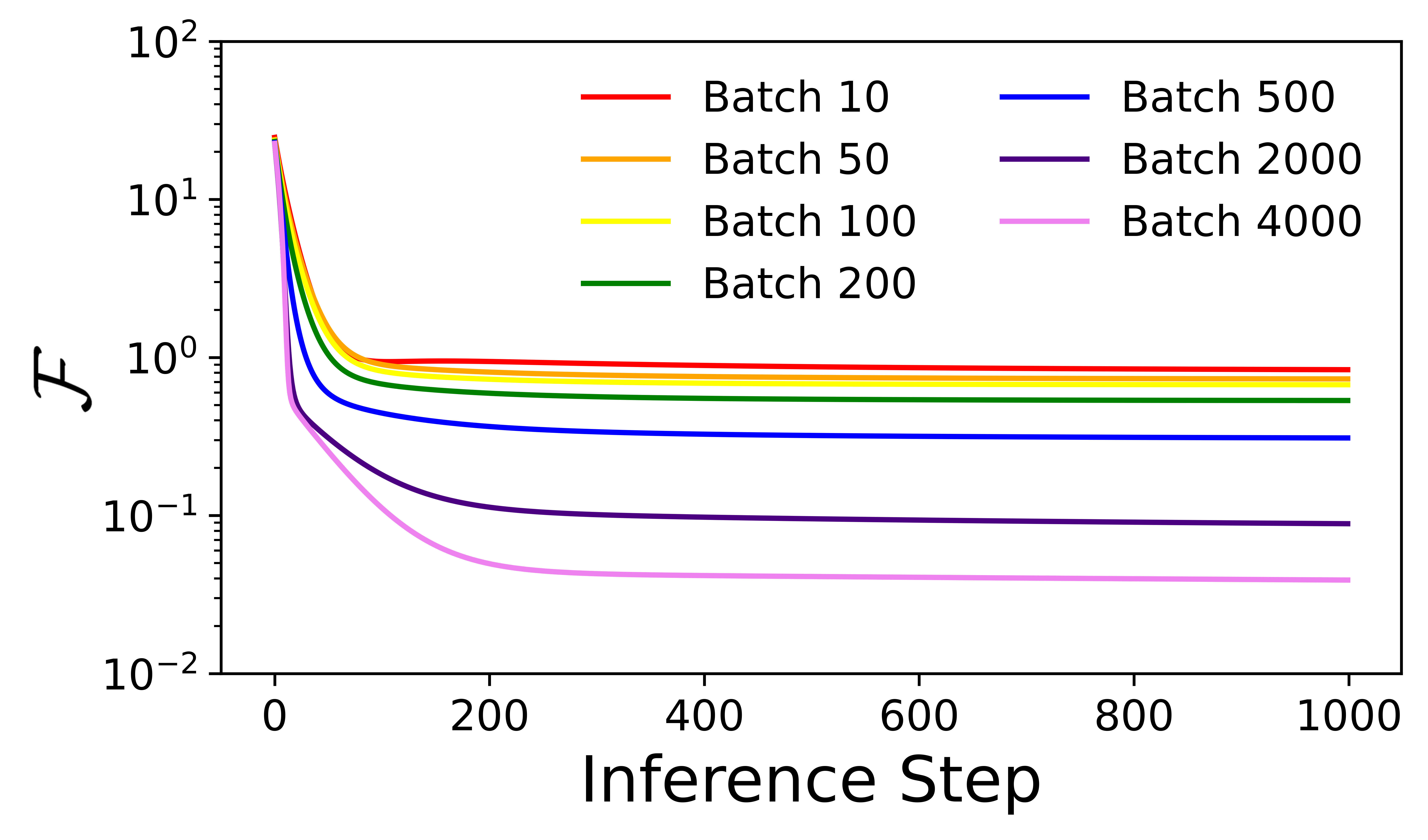}
  \caption{\(\mathcal{F}\) on layer 1}
  \label{F_development:layer1}
\end{subfigure}
\begin{subfigure}{.325\textwidth}
  \centering
  \includegraphics[width=.995\linewidth]{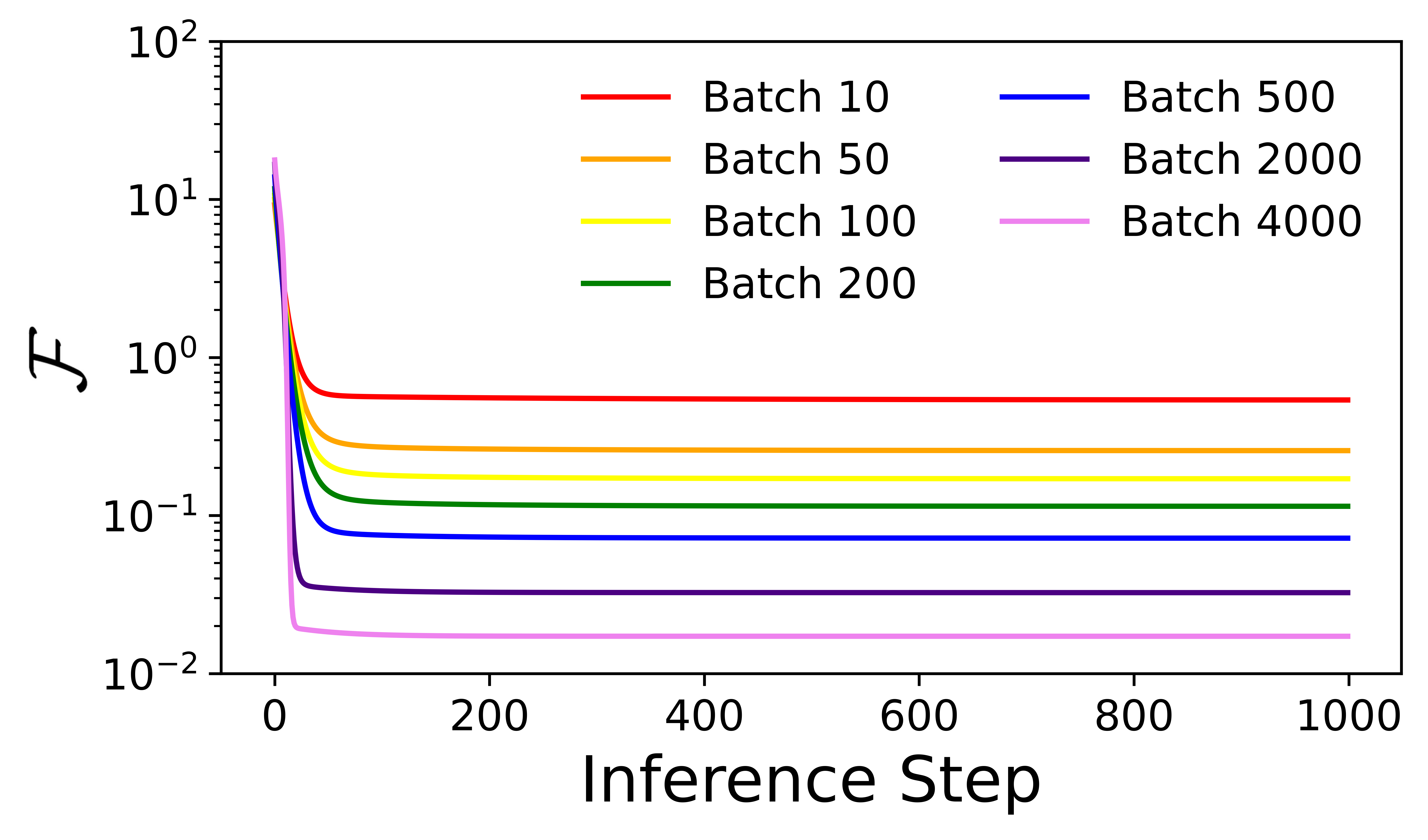}
  \caption{\(\mathcal{F}\) on layer 2}
  \label{F_Development:layer2}
\end{subfigure}
\begin{subfigure}{.325\textwidth}
  \centering
  \includegraphics[width=.995\linewidth]{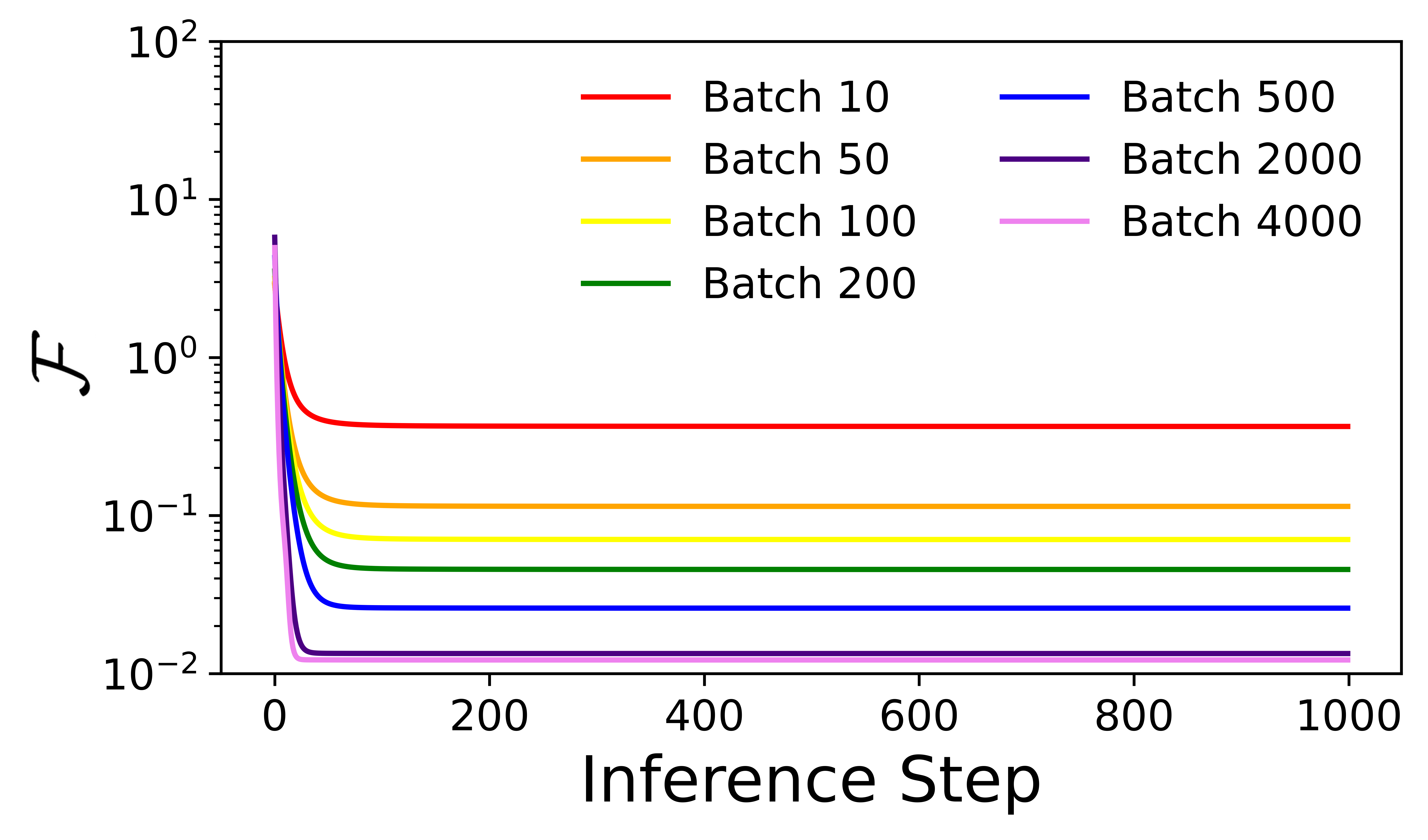}
  \caption{\(\mathcal{F}\) on layer 3}
  \label{F_Development:layer3}
\end{subfigure}
\caption{\textbf{For different levels of training, development of \(\mathcal{F}\) on each layer during inference.} 
Figures show the average \(\mathcal{F}\) per node, for each layer. As training develops, \(\mathcal{F}\) on all layers asymptote at lower values. Layers 2 and 3 also increase the speed which which they stabilise. However, on layer 1 (which represents the difference between the one-hot labels on the top layer and the nodes on the layer below), the network becomes slower to stabilise as training progresses.
\newline This plot shows results from a single run - different runs vary slightly, but follow the same pattern.}
\label{F_development}
\vspace{-0.4cm}
\end{figure}

We have thus discovered the immediate cause of the performance deterioration - as we increase training, we need more test iterations to infer the correct label. To gain a deeper understanding of the phenomenon, however, we need to understand why this is the case.
As an energy based model, the process of inference involves updating nodes until the network reaches equilibrium. Once this has been achieved, \(\frac{d\mathcal{F}}{d\mu}\) for each node will be zero. It is instructive to examine whether any specific part of the network is the cause of the increase in the time to stabilise across epochs. Fig.~\ref{F_development} shows that, as training progresses, the equilibrium \(\mathcal{F}\) for each layer (which is proportional to the mean square error between layers) gets lower. Also, the error nodes on layers 2 and 3 stabilise quickly and even slightly speed up as training progresses. However, \(\mathcal{F}\) on layer 1 (which is a measure of the errors between the node values at layers 0 and 1) takes an increasingly long time to settle down to equilibrium.  Further investigation shows that this is caused by the nodes on the top layer taking increasingly long to stabilise, whereas hidden layer nodes all settle quickly (more details in Appendix~\ref{appendix:nodes}).  

Summarising what this means for our 4 layer network as it tries to infer the label for an image: 
layer 3 at the bottom is fixed with the image,
layer 2 quickly reaches equilibrium with layer 3,
layer 1 quickly reaches equilibrium with layer 2,
layer 0 is not at equilibrium with layer 1 and is still updating after many iterations. 
Crucially, it is this top layer which is used to read out the inferred label - thus causing the apparent reduction in label accuracy during training.

We now examine the reason why it is only the top layer which takes an increasing amount of time to stabilise. Recall that the node values are updated with Euler integration using \(\frac{d\mathcal{F}}{d\mu}\) from equations~\eqref{updates2} and~\eqref{updates3} as the gradient. As we are using fixed precision matrices, we can ignore the precision terms. Also, to help gain an intuition for these gradients, we ignore the non-linear activation function for now and assume the connections are simply linear. This allows us to see that the size of gradients are controlled by sizes of errors and weights:

\begin{equation}
\label{node_update_simple0}
\frac{d\mathcal{F}}{d\mu_0} \ \approx \
\epsilon_{1} \
 \theta_0^T
\end{equation}

\begin{equation}
\label{node_update_simplen}
\frac{d\mathcal{F}}{d\mu_n} \ \approx \
\Big[
\epsilon_{n+1} \
 \theta_n^T \ - \epsilon_{n} \
 \Big]
 \text{ for n} > 0
\end{equation}

\begin{wrapfigure}{r}{0.5\linewidth}
 \vspace{-0.1cm}
 \captionsetup{justification=raggedright}
  \centering
  \includegraphics[width=.9\linewidth]{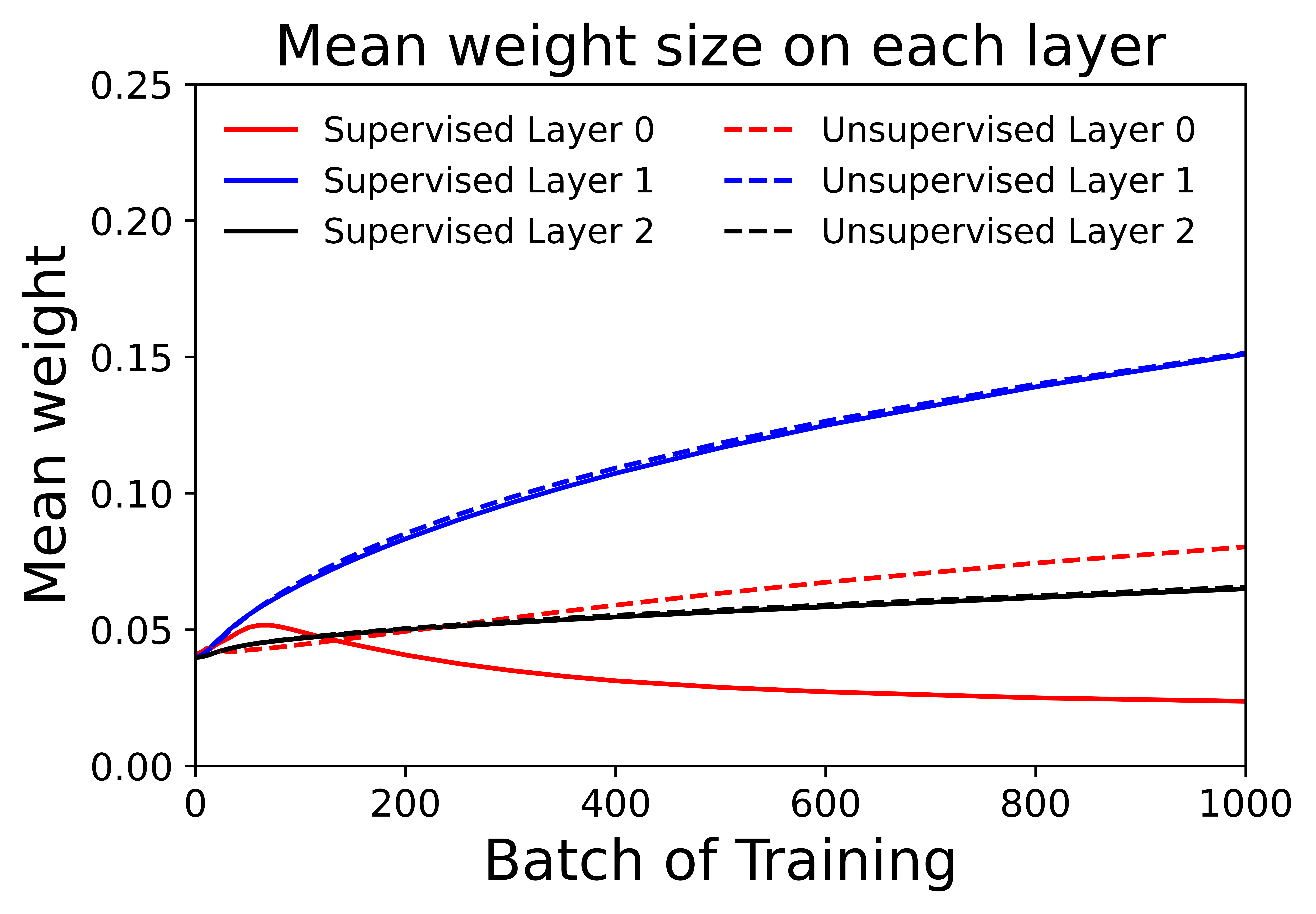}
\caption{\textbf{Development with training of mean weight size for each weight layer.}  Solid lines show supervised mode, with top layer clamped in training - as training progresses, weight size reduces on the top layer but increases on the hidden layers.  On the other hand, if the network is trained in unsupervised mode, all 3 weight layers increase in size.}
\label{weights}
\vspace{-0.1cm}
\end{wrapfigure}

We have seen already that \(\mathcal{F}\), and therefore the errors at all layers, reduces in size as training progresses. 
On the other hand, fig.~\ref{weights} shows that there is a disparity between the weight sizes on the layers.
As training progresses through the batches, the mean weight size reduces on the top layer but increases on the hidden layers. Applying this information to equations~\eqref{node_update_simple0} and~\eqref{node_update_simplen} demonstrates why training causes the gradients on the top layer to reduce, but the hidden layers to increase: the gradient for the top layer is the product of two decreasing quantities, whereas on the hidden layers, diminishing errors are offset by increasing weights.  

This leaves one final level of understanding to be dealt with - why do the weights at the top layer get smaller as training proceeds through the batches but the weights of hidden layers increase? To understand this, we now turn to the update equations for the weights which are applied after each batch of training (equation~\eqref{updates1}. Again, we ignore the precision and non-linearity in order to gain intuition and can see that the update equations for each weight layer \(\theta_n\) comprise:  

\begin{equation}
\label{update_split}
\forall n \ : \    \frac{d\mathcal{F}}{d\theta_n} \ \approx \
\epsilon_{n+1} \ \mu_n^T \ = \ (\mu_{n+1} - \mu_n \theta_n ) \ \mu_n^T \ = \ \mu_{n+1} \ \mu_n ^ T - \mu_n \ \mu_n ^T\ \theta_n \
\end{equation}


This gradient is made up of a Hebbian component and an anti-Hebbian regularisation term and is reminiscent of the Oja learning rule~\cite{Oja:2008} which was proposed as a technique to prevent exploding weights in Hebbian learning. 

Although the equation for the  weight gradient is the same for each layer, there is one crucial difference between the layers: when learning \(\frac{d\mathcal{F}}{d\theta_n}\) in the case  \(n=0\), the nodes \(\mu_n\) are fixed 
, whereas the nodes \(\mu_n\)  are not fixed for other values of \(n\). This interferes with the regularisation of those weight layers. As a result, the weight matrices grow in size when \(n > 0 \).

We have thus seen that accuracy of label inference deteriorates after a certain level of training because the different weight layers are being subject to different regularisation regimes. The top layer nodes are 
fixed during training, keeping the weight updates small, 
and this has the ultimate effect of causing top layer nodes to update more slowly as training progresses. The weights of the hidden layers are updated according to the same equation but with the crucial difference that the input nodes to the weight matrix are not clamped, reducing the anti-Hebbian impact of the regularisation term.  

\section{Techniques to prevent deterioration in inference accuracy}
In order to prevent this deterioration we examine two possible approaches. One approach is to ensure that the top weight layer is treated in the same way as the hidden layers and is not regularised.  Alternatively we can ensure that the hidden layers are regularised in the same way as the top layer. We address these two techniques in turn.

If we unclamp the top layer in training then we are training the PCN in unsupervised mode. Because the top layer of nodes is no longer clamped, the top layer of weights will be treated in the same way as the other layers and the implicit regularisation is reduced. This is demonstrated by the dotted red line in figure~\ref{weights}, which shows that, in unsupervised mode, the top layer weights now continue to increase as training continues just like the middle layers. In addition to monitoring mean weight size of each layer, we can obtain additional intuition as to what is happening by monitoring the development of the singular values of each weight matrix.  This is briefly discussed in Appendix~\ref{appendix:svds}.   

Because training is now unsupervised, the top layer will no longer categorise the MNIST labels according to a one-hot label and therefore we cannot measure speed of label inference as we did in figure~\ref{inference_speed}. \footnote{Note that this does not mean the top layer is no longer useful in terms of representing an image's label - energy minimization will still produce a representation at the top layer which separates the images by their characteristics - see Appendix~\ref{appendix:tsne}}
However, we can analyse the speed with which each layer reduces \(\mathcal{F}\) (as we did in figure~\ref{F_development} for supervised mode). We find that there is no deterioration in time to reach equilibrium in any of the layers (chart not shown) and in fact, like the other hidden layers, the top layer converges slightly quicker as training progresses. We thus find that there is no longer a mismatch between the speeds at which the top layer and the hidden layers reach equilibrium - by having no prior forced on it, the top layer effectively becomes just another hidden layer.

But what if we want to have a supervised network? In that case, we want to maintain inference speed by ensuring that the weights on all layers are similarly regularised. A straightforward way of doing this is to simply ensure that the mean size of the weights for each layer stays constant throughout training. This is done by updating the weights as usual using \(\frac{d\mathcal{F}}{d\theta}\) and then simply regularising them according to the formula

\begin{equation}
\label{normalize}
\forall i \ : \
\theta_i \ =
\ \theta_i * \frac{\text{Target mean size}}{\Sigma_i \lvert \theta_i \rvert},
\end{equation}

where target mean size is a hyperparameter we set as described in Appendix~\ref{appendix:network_details}). The norm of the weight layers are now no longer changing with respect to each other and so, as predicted by our investigations, this prevents deterioration in label inference accuracy - shown by the red line in figure~\ref{diminishing_accy}. A possible challenge to this approach is that this method of regularisation will no longer depend on purely local updates (although there is much neuroscience literature on the topic of homeostatic synaptic scaling - see for example~\cite{turrigiano1998activity,carandini2012normalization}). We used the same regularizing factor on each layer (see Appendix~\ref{appendix:network_details}), and we could find no other configuration of regularizing factor which worked better in terms of classification performance or speed of inference. 

An even simpler, and possibly more biologically plausible, method of preventing exploding weights in supervised learning is to impose a simple cap on each weight.  Empirically, we find that this method also maintains accuracy as shown by the green line in figure~\ref{diminishing_accy}. Results shown were generated using a weight cap size of $0.1$. It should be noted that this method is sensitive to the cap size used (although one could argue that evolution in the brain could select the correct cap size). Also, it may eventually lead to binary weight distribution and declining performance, although we have not investigated this.       

\section{Discussion}

We have provided a detailed analysis of the dynamics of inference in predictive coding, and have shown that there is a tendency for that inference to slow down as training progresses.~\cite{millidge2022backpropagation} separates the total energy of an Energy Based Model into the supervised loss (which depends on the errors at the top layer) and the internal energy (which corresponds to the energy of the hidden layers). We have shown that it is only the supervised loss which suffers from a slow-down in inference.  As a result, this pathology does not exist in unsupervised training. We have also demonstrated that, even in supervised training, the decline can be prevented if the weights are constrained to ensure any weight regularisation is consistently applied across all layers. This is not something that happens automatically in the PCN framework without precisions. \\
\indent In our implementation we have set all precisions to the identity matrix. We are aware of little research on the impact of precisions on inference dynamics, with most mentions of precisions pointing to them as an attention mechanism and a way of controlling the equilibrium point of the network~\cite{feldman2010attention}. But it can be seen from the update equations~\eqref{updates2},~\eqref{updates3} and~\eqref{updates1}, that precisions also act as an adaptive weighting of the learning rate, both in the fast update of nodes and the slower update of weights.  Therefore, as well as influencing direction of gradient updates, they should also have a significant impact on speed of update. Future work therefore needs to address the extent to which the phenomenon we observe is avoided if well-learned precisions are implemented. Having said that, we have experimented with different manually enforced relative precisions at each layer and found little benefit in terms of avoiding inference speed degradation. However, our testing simply implemented constant precisions for each layer - it is entirely possible that, by deriving a full covariance matrix, and allowing precisions to change with time, the phenomenon disappears. If precisions do prove to be a solution, then this paper will have at least pointed out the potential pitfalls of implementing predictive coding without them, and provided some techniques for coping with their omission. \\

\section{Acknowledgements}
\vspace{-0.2cm}
PK is supported by the Sussex Neuroscience 4-year PhD Programme. CLB is supported by BBRSC grant number BB/P022197/1. BM is supervised by Rafal Bogacz who is supported by the BBSRC number BB/S006338/1 and MRC grant number MC\_UU\_00003/1.  
%
%
%
%
%
\bibliographystyle{splncs04}
\bibliography{thebibliography}

\begin{thebibliography}{10}
\providecommand{\url}[1]{\texttt{#1}}
\providecommand{\urlprefix}{URL }
\providecommand{\doi}[1]{https://doi.org/#1}

\bibitem{beal2003variational}
Beal, M.J.: Variational algorithms for approximate Bayesian inference.
  University of London, University College London (United Kingdom) (2003)

\bibitem{bishop2006pattern}
Bishop, C.M., Nasrabadi, N.M.: Pattern recognition and machine learning,
  vol.~4. Springer (2006)

\bibitem{bogacz2017tutorial}
Bogacz, R.: A tutorial on the free-energy framework for modelling perception
  and learning. Journal of mathematical psychology  \textbf{76},  198--211
  (2017)

\bibitem{BuckleyChristopherL2017Tfep}
Buckley, C.L., Chang, S.K., McGregor, S., Seth, A.K.: The free energy principle
  for action and perception: A mathematical review (2017)

\bibitem{carandini2012normalization}
Carandini, M., Heeger, D.J.: Normalization as a canonical neural computation.
  Nature Reviews Neuroscience  \textbf{13}(1),  51--62 (2012)

\bibitem{ClarkAndy2013WnPb}
Clark, A.: Whatever next? predictive brains, situated agents, and the future of
  cognitive science. The Behavioral and brain sciences  \textbf{36}(3),
  181--204 (2013)

\bibitem{dayan1995helmholtz}
Dayan, P., Hinton, G.E., Neal, R.M., Zemel, R.S.: The helmholtz machine. Neural
  computation  \textbf{7}(5),  889--904 (1995)

\bibitem{dempster1977maximum}
Dempster, A.P., Laird, N.M., Rubin, D.B.: Maximum likelihood from incomplete
  data via the em algorithm. Journal of the Royal Statistical Society: Series B
  (Methodological)  \textbf{39}(1),  1--22 (1977)

\bibitem{doya2007bayesian}
Doya, K., Ishii, S., Pouget, A., Rao, R.P.: Bayesian brain: Probabilistic
  approaches to neural coding. MIT press (2007)

\bibitem{feldman2010attention}
Feldman, H., Friston, K.J.: Attention, uncertainty, and free-energy. Frontiers
  in human neuroscience  \textbf{4}, ~215 (2010)

\bibitem{friston2003learning}
Friston, K.: Learning and inference in the brain. Neural Networks
  \textbf{16}(9),  1325--1352 (2003)

\bibitem{friston2005theory}
Friston, K.: A theory of cortical responses. Philosophical transactions of the
  Royal Society B: Biological sciences  \textbf{360}(1456),  815--836 (2005)

\bibitem{friston2008hierarchical}
Friston, K.: Hierarchical models in the brain. PLoS computational biology
  \textbf{4}(11),  e1000211 (2008)

\bibitem{kinghorn2021habitual}
Kinghorn, P.F., Millidge, B., Buckley, C.L.: Habitual and reflective control in
  hierarchical predictive coding. In: Joint European Conference on Machine
  Learning and Knowledge Discovery in Databases. pp. 830--842. Springer (2021)

\bibitem{knill2004bayesian}
Knill, D.C., Pouget, A.: The bayesian brain: the role of uncertainty in neural
  coding and computation. TRENDS in Neurosciences  \textbf{27}(12),  712--719
  (2004)

\bibitem{lecun-mnisthandwrittendigit-2010}
LeCun, Y., Cortes, C.: {MNIST} handwritten digit database  (2010),
  \url{http://yann.lecun.com/exdb/mnist/}

\bibitem{mackay2003information}
MacKay, D.J., Mac~Kay, D.J.: Information theory, inference and learning
  algorithms. Cambridge university press (2003)

\bibitem{millidge2019combining}
Millidge, B.: Combining active inference and hierarchical predictive coding: A
  tutorial introduction and case study. PsyArXiv  (2019)

\bibitem{millidge2021predictive}
Millidge, B., Seth, A., Buckley, C.L.: Predictive coding: a theoretical and
  experimental review. arXiv preprint arXiv:2107.12979  (2021)

\bibitem{millidge2022backpropagation}
Millidge, B., Song, Y., Salvatori, T., Lukasiewicz, T., Bogacz, R.:
  Backpropagation at the infinitesimal inference limit of energy-based models:
  Unifying predictive coding, equilibrium propagation, and contrastive hebbian
  learning. arXiv preprint arXiv:2206.02629  (2022)

\bibitem{millidge2022predictive}
Millidge, B., Tschantz, A., Buckley, C.L.: Predictive coding approximates
  backprop along arbitrary computation graphs. Neural Computation
  \textbf{34}(6),  1329--1368 (2022)

\bibitem{mumford1992computational}
Mumford, D.: On the computational architecture of the neocortex. Biological
  cybernetics  \textbf{66}(3),  241--251 (1992)

\bibitem{Oja:2008}
Oja, E.: {O}ja learning rule. Scholarpedia  \textbf{3}(3), ~3612 (2008).
  \doi{10.4249/scholarpedia.3612}, revision \#91607

\bibitem{rao1999predictive}
Rao, R.P., Ballard, D.H.: Predictive coding in the visual cortex: a functional
  interpretation of some extra-classical receptive-field effects. Nature
  neuroscience  \textbf{2}(1),  79--87 (1999)

\bibitem{salvatori2022learning}
Salvatori, T., Pinchetti, L., Millidge, B., Song, Y., Bogacz, R., Lukasiewicz,
  T.: Learning on arbitrary graph topologies via predictive coding. arXiv
  preprint arXiv:2201.13180  (2022)

\bibitem{seth2014cybernetic}
Seth, A.K.: The cybernetic bayesian brain. In: Open mind. Open MIND. Frankfurt
  am Main: MIND Group (2014)

\bibitem{song2020can}
Song, Y., Lukasiewicz, T., Xu, Z., Bogacz, R.: Can the brain do
  backpropagation?---exact implementation of backpropagation in predictive
  coding networks. Advances in neural information processing systems
  \textbf{33},  22566--22579 (2020)

\bibitem{tschantz2022hybrid}
Tschantz, A., Millidge, B., Seth, A.K., Buckley, C.L.: Hybrid predictive
  coding: Inferring, fast and slow. arXiv preprint arXiv:2204.02169  (2022)

\bibitem{turrigiano1998activity}
Turrigiano, G.G., Leslie, K.R., Desai, N.S., Rutherford, L.C., Nelson, S.B.:
  Activity-dependent scaling of quantal amplitude in neocortical neurons.
  Nature  \textbf{391}(6670),  892--896 (1998)

\end{thebibliography}
\vfill

\pagebreak
\appendix
\section{Network Details}
\label{appendix:network_details}

 \textbf{Network size}: 4 layer

\noindent  \textbf{Number of nodes on each layer}: 10, 100, 300, 784. In both training and testing, the 784 bottom layer nodes were fixed to the MNIST image. In supervised training mode, the top layer nodes were fixed as a one-hot representation of the MNIST label. 

\noindent  \textbf{Non-linear function}: tanh

\noindent  \textbf{Bias used}: yes

\noindent  \textbf{Training set size}: full MNIST training set of 60,000 images, in batches of 640. Thus the full training set is used after 93 batches.

\noindent  \textbf{Testing set size}: full MNIST test set of 10,000 images

\noindent  \textbf{Learning parameters used in weight update of EM process}: Learning Rate=  1e-3, Adam

\noindent  \textbf{Learning parameters used in node update of EM process}: Learning Rate=  0.025, SGD

\noindent  \textbf{Number of SGD iterations in training}: 50

\noindent  \textbf{Number of SGD iterations in testing}: 1000.

\noindent  \textbf{Random node initialisation}:  Except where fixed, all nodes were initialized with a random values selected from \(\mathcal{N}(0.5, 0.05)\)

\noindent
\textbf{Weight regularisation}: The weight regularisation technique used holds the L1 norm of each weight matrix constant. Rather than assigning a specific value, the algorithm measured the L1 norm of the matrix at initialisation (before any training took place) and then maintained that norm after each set of weight updates. The weights were randomly initialised using \(\mathcal{N}(0,0.05)\), giving a mean individual weight size of approximately 0.04 on all layers.

\section{Development of nodes}
\label{appendix:nodes}

\begin{figure}
 \vspace{-.4cm}
 \captionsetup{justification=raggedright}
\centering
\begin{subfigure}{.325\textwidth}
  \centering
  \includegraphics[width=.995\linewidth]{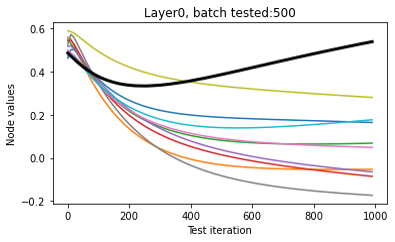}
  \caption{After 500 batches of training \newline}
  \label{nodes_supervised:1}
\end{subfigure}
\begin{subfigure}{.325\textwidth}
  \centering
  \includegraphics[width=.995\linewidth]{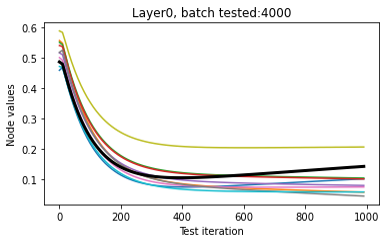}
  \caption{After 4000 batches of training \newline}
  \label{nodes_supervised:2}
\end{subfigure}
\begin{subfigure}{.325\textwidth}
  \centering
  \includegraphics[width=.995\linewidth]{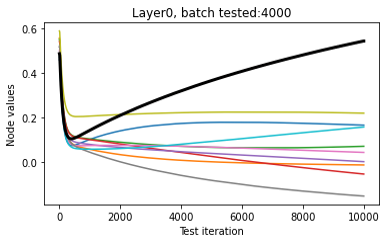}
  \caption{After 4000 batches of training - with more test iterations}
  \label{nodes_supervised:3}
\end{subfigure}
\caption{\textbf{Nodes on top layer take longer to approach equilibrium after many epochs of training.}  For a PCN trained in supervised mode, the figures show how the 10 one-hot nodes in the top layer develop during test inference. The black lines represent the node which corresponds with the presented MNIST image's label (and therefore should be the maximum of the 10 nodes).  Figure (a) shows the situation relatively early in training; after less than the 1000 test iterations, the system is inferring the correct label. But figure (b) shows the situation when training has run much longer; the nodes are now much slower to update and therefore the system infers the wrong label.  However, if the system were allowed to carry out inference over more iterations, the correct label would be inferred (figure c). }
\label{nodes_supervised}
\vspace{-0.4cm}
\end{figure}
\vfill

\pagebreak
\section{Development of SVDs}
\label{appendix:svds}
\begin{figure}
\vspace{-.4cm}
 \captionsetup{justification=raggedright}
\centering

\begin{subfigure}{.3\textwidth}
  \centering
  \includegraphics[width=.995\linewidth]{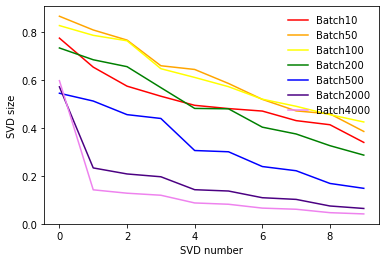}
  \caption{No weight constraints - layer 0}
  \label{SVD_dev:layer0}
\end{subfigure}
\begin{subfigure}{.3\textwidth}
  \centering
  \includegraphics[width=.995\linewidth]{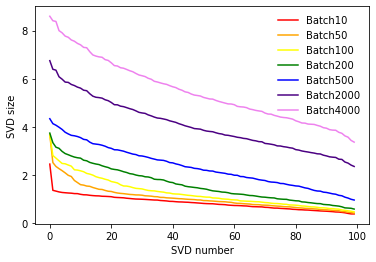}
  \caption{No weight constraints - layer 1}
  \label{SVD_dev:layer1}
\end{subfigure}
\begin{subfigure}{.3\textwidth}
  \centering
  \includegraphics[width=.995\linewidth]{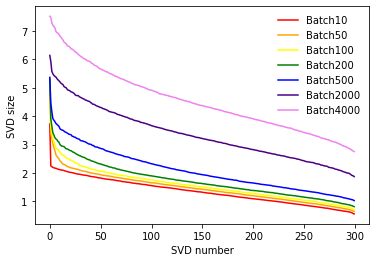}
  \caption{No weight constraints - layer 2}
  \label{SVD_dev:layer2}
\end{subfigure}

\begin{subfigure}{.3\textwidth}
  \centering
  \includegraphics[width=.995\linewidth]{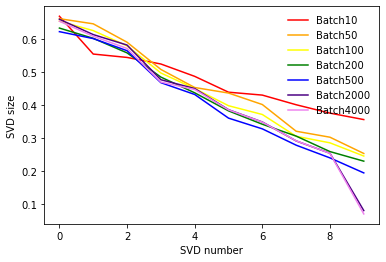}
  \caption{Normalized weights - layer0}
  \label{SVD_dev:Normlayer0}
\end{subfigure}
\begin{subfigure}{.3\textwidth}
  \centering
  \includegraphics[width=.995\linewidth]{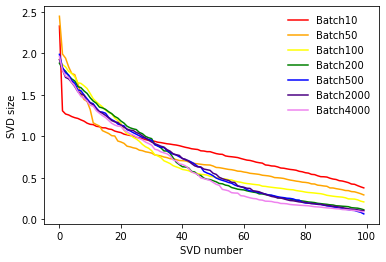}
  \caption{Normalized weights - layer1}
  \label{SVD_dev:Normlayer1}
\end{subfigure}
\begin{subfigure}{.3\textwidth}
  \centering
  \includegraphics[width=.995\linewidth]{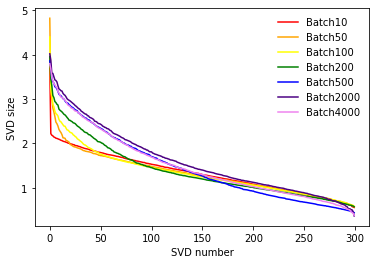}
  \caption{Normalized weights - layer2}
  \label{SVD_dev:Normlayer2}
\end{subfigure}

\begin{subfigure}{.3\textwidth}
  \centering
  \includegraphics[width=.995\linewidth]{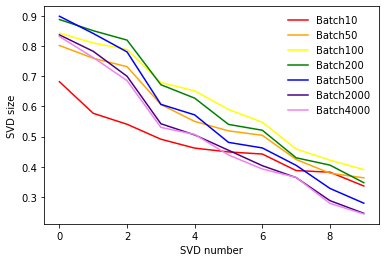}
  \caption{Capped weights - layer 0}
  \label{SVD_dev:Caplayer0}
\end{subfigure}
\begin{subfigure}{.3\textwidth}
  \centering
  \includegraphics[width=.995\linewidth]{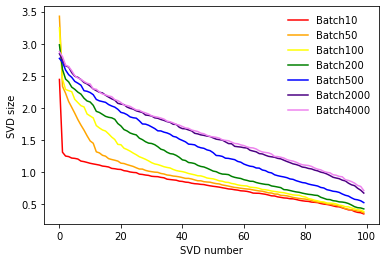}
  \caption{Capped weights - layer 1}
  \label{SVD_dev:Caplayer1}
\end{subfigure}
\begin{subfigure}{.3\textwidth}
  \centering
  \includegraphics[width=.995\linewidth]{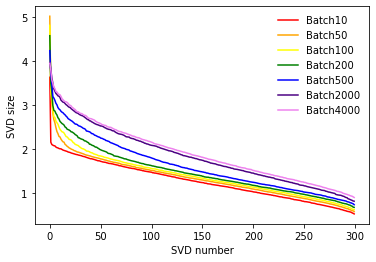}
  \caption{Capped weights - layer 2}
  \label{SVD_dev:Caplayer2}
\end{subfigure}

\caption{\textbf{Singular value decomposition of weight matrices.} If we do not constrain the weights (figures a - c), the SVDs of layer 0 increase in early stages of training, and then decrease. But layers 1 and 2 steadily increase. In early stages of training, the distribution of SVDs changes, but after peak inference accuracy has been achieved at around training batch 500, subsequent changes in the weight matrix can be viewed largely as a parallel increase in all SVDs. \\ By introducing normalisation (figures d - f), the network learning is still able to redistribute the shape of the weight matrix if the gradients try to do this. But if the gradients are having no effect on the shape of the matrix then changes are not applied. Similar behaviour is observed for capped weights (figures g  - i). \\
Monitoring the shape of SVDs could help identify when there is nothing to be gained from further training, although this would probably also require monitoring the change in singular vectors.}
\label{SVD_dev}
\vspace{-0.4cm}
\end{figure}
\vfill

\pagebreak
\section{Unsupervised PCN still separates top layer into labels}
\label{appendix:tsne}
\begin{figure}
 \vspace{-.4cm}
 \captionsetup{justification=raggedright}
  \centering
  \includegraphics[width=.9\linewidth]{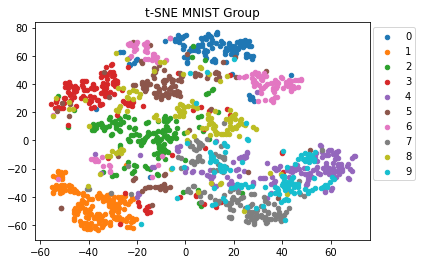}
\caption{\textbf{Unsupervised PCN.} Because the network is trained in an unsupervised manner, the top layer nodes do not contain a one-hot estimate of the image label. The nodes still contain a representation of the image, but it is simply in a basis that the network has created, rather than one which has been forced on it using supervised learning. tSNE analysis of top layer in testing demonstrates that unsupervised PCN still separates images according to label, despite lack of labels in training. Each dot shows the tSNE representation for an MNIST image - different colours represent images with different labels. }
\label{tsne}
\vspace{-0.4cm}
\end{figure}

\end{document}